\documentclass{article}

\PassOptionsToPackage{numbers, compress}{natbib}
\usepackage[final]{neurips_2022}

\usepackage[utf8]{inputenc} % allow utf-8 input
\usepackage{hyperref}       % hyperlinks
\usepackage{url}            % simple URL typesetting
\usepackage{booktabs}       % professional-quality tables
\usepackage{amsfonts}       % blackboard math symbols
\usepackage{nicefrac}       % compact symbols for 1/2, etc.
\usepackage{microtype}      % microtypography
\usepackage{xcolor}         % colors
\usepackage{booktabs,siunitx}
\usepackage[english]{babel}
\usepackage[utf8]{inputenc}
\usepackage{algorithm} 
\usepackage[noend]{algpseudocode}
\usepackage{multirow}
\usepackage{graphicx}
\usepackage{wrapfig}
\usepackage{subfigure} 
\usepackage{threeparttable}
\newcommand{\yonggang}[1]{\textcolor{black}{#1}} % stands for new content added by yonggang
\newcommand{\chsun}[1]{\textcolor{black}{#1}}
\usepackage{ulem}
\newtheorem{theorem}{Theorem}
\newtheorem{definition}[theorem]{Definition}

\title{Towards Lightweight Black-Box Attacks against Deep Neural Networks}

\author{
Chenghao Sun$^{1}$ \quad Yonggang Zhang$^{2}$ \quad  Wan Chaoqun$^3$ \quad Qizhou Wang$^2$ \quad
\\\bf Ya Li$^4$ \quad   Tongliang Liu$^5$ \quad   
 Bo Han$^{2}$ \quad  
\bf Xinmei Tian$^{1\thanks{  Corresponding author (xinmei@ustc.edu.cn)}}$
\quad  \\
 $^1$University of Science and Technology of China \quad
 $^2$Hong Kong Baptist University \\ 
 $^3$Alibaba Cloud Computing Ltd \quad
 $^4$iFlytek Research \\
 $^5$The University of Sydney \\ 
 \setcounter{footnote}{0}
}

\begin{document}

\maketitle

\begin{abstract}
Black-box attacks can generate adversarial examples without accessing the parameters of deep neural networks (DNNs), largely exacerbating the threats of deployed models. However, previous works state that black-box attacks fail to mislead DNNs when their training data and outputs are inaccessible. 
In this work, we argue that black-box attacks can pose practical attacks in this highly restrictive scenario where only several test samples are available. 
Specifically, we find that attacking the shallow layers of DNNs trained on a few test samples can generate powerful adversarial examples. As only a few samples are required, we refer to these attacks as \textit{lightweight} black-box attacks.
The main challenge to promoting lightweight attacks is to mitigate the adverse impact caused by the approximation error of shallow layers. 
As it is hard to mitigate the approximation error with few available samples, we propose Error TransFormer (ETF) for lightweight attacks.
Namely, ETF transforms the approximation error in the parameter space into a perturbation in the feature space and alleviates the error by disturbing features. In our experiments, lightweight black-box attacks with the proposed ETF achieve surprising results. For example, even if only $1$ sample per category is available, the attack success rate achieved by lightweight black-box attacks is only about $3\%$ lower than that of the black-box attacks using complete training data~\footnote{Code is available at \url{https://github.com/sunch-ustc/Error_TransFormer/tree/ETF}}.

\end{abstract}

\section{Introduction}
\label{sec:Introduction}
Black-box attack methods~\cite{TI-FSGM,DI-FSGM,black-box-cite} can mount successful attacks without accessing the parameters of deep neural networks (DNNs), posing great challenges to deep learning in safety-critical situations. Existing black-box attack methods implicitly assume that the training data and/or the outputs of target models are available to adversaries. For scenarios where training data are available, adversarial examples can be generated by attacking surrogate models constructed for approximate target models. Using the target models' outputs for estimating gradients is another practical approach to crafting adversarial examples. In general, due to the low dependency on target model information, the black-box attacks raise realistic threats to real-world deep learning systems, including semantic segmentation~\cite{cite35,cite36}, object detection~\cite{cite37,cite38,cite39}, and automatic driving~\cite{auto}. 

However, these assumptions can be violated in many practical scenarios~\citep{ZOO,survey}, where both the training data and the outputs of target models are inaccessible. This realistic attack scenario is introduced in~\citep{ZOO}, generally termed as the \textit{no-box setting}. Under such a restrictive setting, Li et al.~\cite{nobox} stated that existing black-box attacks fail to mislead target models because it is now forbidden to access previously essential information in mounting black-box attacks. Therein, the failure of existing black-box attacks is not surprising. The reason is that constructing reliable surrogate models typically relies on accessing the complete training data of target models, which is forbidden in such a restrictive setting with only a few test samples available. 

In this work, we aim to reveal the potential threats of black-box attacks in the no-box setting, as mounting black-box attacks with highly limited accessible information leaves many models exposed to attack. Specifically, we challenge the previous believes by raising the following question: {\textit{Can black-box attacks success in the no-box setting?}} The doubts about the potential threats of the black-box attack are not groundless. Specifically, the potential threats of black-box attacks stem from two facts: a) adversarial examples can be generated by perturbing representations at shallow layers of DNNs~\cite{MSE-LOW,FDA_2020NIPS,Intermediate_layer_ICLR2016}; b) regarding the representation of shallow layers, there do not exist critical differences between those models learned from a few data and that of the whole training data.~\cite{low-level}

Building upon the above facts, it is actually possible that an adversary can successfully mount attacks using limited samples, as powerful adversarial examples can be generated by attacking the shallow layers of DNNs trained on few samples. Concretely, the adversary can leverage available samples to construct a surrogate model to approximate the shallow layers of target models within acceptable errors. Consequently, adversarial examples can be generated by perturbing the features obtained from the shallow layers of the constructed surrogate model. As merely a few samples are required for attacking, we refer to black-box attacks in the no-box setting as \textit{lightweight black-box attacks}. Intuitively, the closer the surrogate model is to the target model, the higher the attack success rate the crafted adversarial examples have~\citep{scale}. If the approximation error of shallow layers is alleviated, the lightweight black-box attack can be as powerful as black-box attacks with complete training data. Hence, the main challenge to mounting a lightweight attack is to mitigate the adverse impact caused by the approximation error of shallow layers.

However, it is challenging to mitigate the approximation error, especially when the number of available samples is limited. Fortunately, it is straightforward to identify which kind of perturbations are preferred: if perturbations applied to a feature contribute to fooling the surrogate model, the perturbation is preferred. Therefore, bridging the connection between the parameter space and the feature space is the key to mitigating the approximation error. Accordingly, we propose transforming the approximation error in the parameter space as the perturbation in the feature space, dubbed Error TransFormer (ETF). Namely, ETF transforms the worst-case approximation error to the worst-case feature perturbation, leading to a min-max scheme that generates adversarial examples under the worst feature perturbations. We verify the attack success rate of lightweight black-box attacks using $7$ models trained on the ImageNet dataset~\cite{ImageNet} and find that existing attack methods are much more potent than previously claimed. Moreover, the performance of lightweight black-box attacks can be further promoted by the proposed ETF, i.e., the attack success rate is only $3\%$ lower than that achieved by black-box attacks having complete training data of target models.

\section{Related works}
\label{sec:Related Work}
% DI MI TI related work:https://papers.nips.cc/paper/2021/file/30d454f09b771b9f65e3eaf6e00fa7bd-Paper.pdf
\subsection{Adversarial attack}
\label{subsec:Related Work-adversarial attacks}
According to the amount of accessible information, existing adversarial attacks can be roughly divided into two categories: white-box attacks and black-box attacks.
White-box attacks~\cite{FGSM,deepfool,black-mmd,Robustness-cite} assume that the target model is transparent to adversaries, i.e., adversaries can access all information about the target model. Nevertheless, the assumption  of transparent target models can be violated in many practical scenarios~\cite{MI-FGSM}. 
Hence, black-box attacks~\cite{black-box-cite3,black-box-cite2,black-gang,black-cite4} are proposed and applied to the scenario where relatively limited information about the target model is accessible, i.e., only a certain number of model queries or the training data of target tasks are available.    
%Utilizing the training data or model queries, adversaries generate adversarial examples on the surrogate model constructed for approximating target models. 

% Intermediate level attacks------
Among these black-box attacks, intermediate-level attacks~\citep{Intermediate_layer_ICLR2016,MSE-LOW,ILA} are widely explored to improve the adversarial transferability. The core idea of these attacks is based on the empirical observation that well-trained models' intermediate features are transferable~\citep{low_level-first_few_layers}. This is consistent with the recent works showing that adversarial examples comprise spurious features having the transferable property~\cite{causal-gang,chen2020self}. Hence, Inkawhich et al.~\cite{MSE-LOW} propose to perturb the feature space of neural networks to create more transferable adversarial examples. These methods provide an interesting empirical conclusion that disturbing the shallow layers of DNNs can also generate adversarial examples. 

Existing black-box attacks assume that the large-scale training data of target models and the feedback from querying target models are accessible, but these assumptions can be violated in many scenarios, e.g., the no-box setting~\citep{ZOO}. Li et al.~\cite{nobox} stated that existing black-box attacks cannot be successfully mounted because existing DNNs require large-scale training data for generalization. Given only small-size data available, obtaining a surrogate model with strong generalization is challenging. Hence, Li et al. propose replacing black-box attacks with their proposed no-box attack~\cite{nobox}, where they train a classical auto-encoder model instead of the supervised classification model due to the constraint of a small-scale dataset. Concretely, they train $20$ auto-encoders for each category and generate adversarial examples by attacking these auto-encoders, which is time- and computational-consuming.

\yonggang{Different from existing black-box attacks, lightweight black-box attacks aim to reveal the potential risk of black-box attacks under the no-box setting. Because one main challenge to perform lightweight black-box attack is to mitigate the adverse impact caused by the approximation error, we propose error transformer that is a min-max strategy transforming the approximation error in the weight space to the feature space. The min-max strategy is different from that introduced in~\citep{awp}, where min-max strategy is proposed to flatten the loss landscape in the weight space, see Appendix~\ref{perturbation}.}

\subsection{Approximation of Shallow Networks}
A recent study~\cite{low_level-first_few_layers} confirms the intuition that shallow layers in DNNs can be seen as low-level feature extractors, provided that strong data augmentation is used~\cite{low-level}. Motivated by this observation, Asano et al.~\cite{low-level} design a method to explore the information in every layer of DNNs. Asano et al.~\cite{low-level} train a supervision model on ImageNet LSVRC-12, and a self-supervision model on a small-scale dataset. Then, they apply linear probes~\cite{linear_probes} to all intermediate layers of networks, where a linear classifier is trained on the top of pre-trained and fixed feature representations. In this way, they evaluate the quality of the representation learned at different depths of the networks. The results show that, given heavily synthetic transformations, the shallow layers of DNNs learned from a few images can approximate that of DNNs trained on millions of images. According to the intriguing empirical observation, it is possible to construct a surrogate model with limited samples, where its shallow layers are similar to that of the target models. Besides, it is known that the neural networks prefer to make the decision through the spurious feature~\cite{causal-gang} on which are focused by models to correlate to the true label. Since shallow layers based on the small-scale or large-scale data set can acquire the similar spurious feature~\cite{low-level}, it is an effective way to leverage the shallow model to mount an attack in the spurious features when the number of the data is limited.

\section{Preliminaries}
\label{sec:Preliminaries}
%This section defines the no-box threat model and the lightweight black-box attack.

\subsection{No-Box Threat Model}
The no-box setting~\cite{ZOO} denotes a threat model where available information about the target model is extremely limited, making it challenging to generate adversarial examples. According to the description in ~\cite{ZOO}, the no-box threat model can be defined as follows.
\begin{definition}[No-Box Threat Model]
A threat model is called the no-box threat model if the adversary in this scenario is not allowed to access the target model's training data and outputs, so it only has some samples that can be correctly predicted with high probability by the target model.
\end{definition}
The definition is consistent with the no-box setting introduced in~\citep{ZOO}, where the target model's training data and outputs are inaccessible to the adversary. Moreover, the available samples used for generating adversarial examples are expected to be limited. This is because if the number of samples is similar to or even larger than the number of samples used for training target models, the adversary can leverage these samples to construct a surrogate model similar to target models, making the condition of inaccessible training data meaningless. It is intuitive that target models can correctly predict the labels of these available samples with high probability. For example, it is meaningless to perform attacks using samples misclassified by the target model. 

%\textbf{Problem Statement .}
%In the no-box attack setting, the adversaries can only gather a small-scale dataset from the same problem domain as that of the target model. The previous study on no-box attacks~\cite{nobox} assume that adversary can collect about twenty images in each category. To explore potential threats of the black-box attack, we assume that only one sample per category is available for adversaries in this paper.

%in the different real world, both the category of collected samples in the data and the number of collected samples in each category are often uncertain.In some extreme conditions, some categories may have only a single image and even the label information is  also difficult to obtain due to the privacy consideration~\cite{survey}. 

\subsection{Lightweight Black-Box Attack}
\begin{definition}[Lightweight Black-Box Attack]
An attack is called the lightweight black-box attack if an adversary aims to perform an attack in the context of a no-box threat model.
\end{definition}
The definition of the lightweight black-box attack shows that lightweight attack is a special kind of black-box attacks. The extremely limited accessible information distinguishes it from other black-box attacks. For example, query-efficient black-box attacks require the feedback information of target models~\citep{nes,pc}, while lightweight attacks do not. 
%Intuitively, many black-box attacks can become the lightweight black-box attacks when the utilized samples are limited, e.g., the translation invariance (transfer) attack~\citep{TI-FSGM}. 
To our best knowledge, how to perform a lightweight black-box attack is still lacking in the literature. For example, Li et al.~\citep{nobox} claim that black-box attacks fail to fool target models when their training data and outputs are inaccessible.

% In this paper, we define the surrogate model trained in the no-box setting as the lightweight surrogate mode. Similarly, we define the surrogate model trained on the training data the general surrogate mode.

% \textbf{Notation.}
%参考这篇文章 https://papers.nips.cc/paper/2021/file/103303dd56a731e377d01f6a37badae3-Paper.pdf
% We define the small-scale dataset $D$. We denote the source images and the guide images as $x_{source}$ and $x_{guide}$, which are from the collected small-scale dataset. $x_{guide}$ and $x_{source}$ come from different categories. The transformation of image $x$ is define as $T(x)$. Image $x$ and its augmentation $\{x,T_{1}(x),T_{2}(x),T_{3}(x)...T_{N}(x)\}$ is denoted as $A_{N}(x)$.

\section{Approach}
\label{sec:Approach}
This section gives a detailed description of how to perform lightweight black-box attacks and the proposed Error TransFormer (ETF) to alleviate to adverse impact caused by approximation error. 

%Generating adversarial examples in the lightweight black-box attacks requires constructing surrogate models. As the training procedure requires limited samples, we refer to these models as lightweight surrogate models. After constructing the surrogate model, we can generate adversarial examples by attacking the model in the feature space, i.e., feature space perturbation. The reason is that using limited samples can merely approximate the shallow layers within acceptable errors. To further promote the performance of lightweight attacks, we propose the Error TransFormer (ETF) to alleviate the adverse impact caused by approximation error.

\subsection{Lightweight Surrogate Model}
Surrogate models are widely used in black-box attacks since adversarial examples are usually generated by attacking surrogate models when querying target models is forbidden. Since the lightweight black-box attack is a special kind of black-box attack, performing the attack also requires constructing surrogate models. We call surrogate models employed in lightweight black-box attacks the \textit{lightweight surrogate model}, building upon the fact that the number of available samples in the no-box threat model is typically limited.

Similar to existing black-box attacks, we assume that the label information of samples used for generating adversarial examples are accessible. In this scenario, the lightweight surrogate model is trained for a classification task. Specifically, the lightweight model is realized by training a classification DNN equipped via conventional supervised learning, using cross-entropy loss $\ell(\cdot, \cdot)$:
\begin{equation} \label{eq:train1}
\min_{w} \mathbb{E}_{(x, y) \sim \hat{\mathcal{D}}} \ \ell(f(x;w), y),
\end{equation}
where $w$ stands for the parameters of the surrogate model $f$, and $(x, y)$ denotes the sample label pair of random variables sampled from the distribution of natural data $\hat{\mathcal{D}}$. 

In black-box attacks, adversary only leverages the shallow layers of DNNs, leading to a straightforward approach to constructing lightweight surrogate models. Specifically, we can train the lightweight surrogate model in a contrastive manner, where the supervised information is no longer necessary:
\begin{equation} \label{eq:train2}
\min_{w} \ \frac{1}{|\hat{D}|} \sum_{\mathbf{x} \in \hat{D}} 
- \log 
\frac{\sigma(f(\mathbf{x};w), f(\mathcal{T}(\mathbf{x});w))}
{\sum_{\mathbf{x}' \sim \hat{D} \backslash \mathbf{x}} \ \sigma(f(\mathbf{x};w), f(\mathcal{T}(\mathbf{x}');w)) / (|\hat{D} \backslash \mathbf{x}| - 1)},
\end{equation}
where $\sigma$ is a specific similarity metric, e.g., cosine similarity, $\mathcal{T}$ denotes the data transformation operation, and $\hat{D} \backslash \mathbf{x}$ means that $\mathbf{x}$ is removed from the dataset $\hat{D}$. The contrastive strategy in Eq.~\ref{eq:train2} makes it possible to perform attacks when the label information is unavailable. \yonggang{Unless otherwise specified, we mainly uses Eq.~\ref{eq:train1} to train the surrogate models.}

\subsection{Feature Space Perturbation}
Built upon the empirical observation~\citep{low_level-first_few_layers}, using limited samples can approximate the shallow layers of DNNs within acceptable errors. This suggests that deep layers other than shallow layers can cause large approximation error. Accordingly, merely the shallow layers of the learned lightweight surrogate models are used for generating adversarial examples, which is similar to the feature space attacks~\citep{Intermediate_layer_ICLR2016}. For brevity, we define the shallow layers as $\varphi$ and denote the feature as $\varphi(x;w)$.

In feature space attacks, guide images are usually required for generating adversarial examples~\citep{Intermediate_layer_ICLR2016}. A natural image used for generating adversarial example is usually called source image. Given a source image $x_{s}$, we generate an adversarial example $x_{adv}$ by perturbing $x_{s}$ such that its feature is similar to that of a guide image $x_{g}$, where the labels of these two images are different. Specifically, the adversarial examples are generated as follows:
\begin{equation} \label{d_metric}
    x_{adv} = \arg \min_{\left \| x' - x\right \|_{p} \leq \epsilon} 
    d( \varphi(x_g; w), \varphi(x'; w) ),
\end{equation}
where $d$ stands for a specific distance, $\epsilon$ controls the strength of adversarial perturbation, and $x'$ denotes the perturbed version of the source image.

\subsection{Error Transformer}
Although applying feature space attacks to lightweight surrogate models can generate transferable adversarial examples, the approximation error will cause adverse impact to mounting attacks. The reason is that large dissimilarity between the surrogate models and the target models reduces the attack success rate. Therefore, the main challenge to mounting lightweight attacks is to mitigate the adverse impact caused by the approximation error. However, available samples are usually limited in the no-box threat model, making it hard to alleviate the approximation error.

\yonggang{To address the challenge, we propose transforming the parameter space's approximation error as the feature space's perturbation. Specifically, we seldom know which perturbations can point (from the surrogate model) to the target model, making it challenging to alleviate the approximation error in the weight space. In contrast, we have the prior that samples with different labels should have distinguishable representations. Thus, we can leverage such prior knowledge to select preferred perturbations in the feature space, i.e., we prefer perturbations that can make representations of samples with different labels indistinguishable. Similarly, it is also straightforward to define the ``bad'' perturbations, leading to the design of the min-max optimization to identify the ``worst'' model.}
Thus, connecting the parameter space and feature space makes it possible to mitigate the adverse impact caused by approximation error. To bridge the connection, we introduce the following key identity, which is also used in~\citep{transform}:
\begin{small}
\begin{equation} \label{eq:trans}
\varphi(x; \left \{w^1 + w^1 A \right \} \cup \left \{ w \backslash w^1 \right \}) = 
g((w^1 + w^1 A) x; w \backslash w^1) =
g(w^1(x+Ax); w \backslash w^1) =
\varphi(x + A x ; w),
\end{equation} 
\end{small}
\hspace{-1mm}where $w^1$ stands for the first layer parameters of model $\varphi$, $A$ is a transformation matrix used for perturbing $w^1$, $\left \{w^1 + w^1 A \right \} \cup \left \{ w \backslash w^1 \right \}$ means that the first layer's parameters are perturbed, the other layers' parameters keep unchanged, and $g$ is the function parameterized with $w \backslash w^1$ used for processing the first layer's outputs. According to Eq.~\ref{eq:trans}, we can transform a perturbation in the parameter space as the perturbation in the data space, i.e., $\varphi(x; \left \{w^1 + w^1 A \right \} \cup \left \{ w \backslash w^1 \right \}) = \varphi(x + A x ; w)$. Similarly, we can apply the transformation operation to the hidden layers, see Appendix~\ref{hiddern} for more details.

Built upon the connection in Eq.~\ref{eq:trans}, we can transform the first layer parameters' discrepancy between the target model and the surrogate model. Taking the first layer as an example, let $w_t$ be the parameters of the target model and $w^1_t$ be its first layer parameters, where the other parameters are the same as the surrogate model. Assume~\footnote{\yonggang{Discussion about the assumption can be found in Appendix~\ref{assumption}.}} that there exists a transformation matrix $A$ applied to the first layer parameters of the surrogate model such that $w^1_t = w^1 + w^1 A$, we have:
\begin{small}
\begin{equation} \label{eq:trans2}
\varphi(x; w^1_t \cup \left \{ w_t \backslash w^1_t \right \}) = 
\varphi(x; \left \{w^1 + w^1 A \right \} \cup \left \{ w_t \backslash w^1_t \right \})
=
\varphi(x + Ax; \left \{w^1 \right \} \cup \left \{ w_t \backslash w^1_t \right \})
=
\varphi(x + Ax; w).
\end{equation}  
\end{small}
\hspace{-1mm}The identity in Eq.~\ref{eq:trans2} shows that we can find a perturbation in the data space to alleviate the difference in the first layer parameters between these two models. Built upon the transformation, we can transform the worst approximation error to the worst perturbation in the data space. Therefore, we can employ a min-max strategy to mitigate the adverse impact of approximation error through generating adversarial examples under the worst perturbation. Thus, lightweight black-box attacks with ETF generate adversarial examples as follows:
\begin{equation}
x_{adv} = \arg 
    \min_{\left \| x' - x\right \|_{p} \leq \epsilon} \  \max_{\left \| \Delta_s \right \|_{p} \leq \tau, \ 
    \left \| \Delta_g \right \|_{p} \leq \tau} \ d(\varphi(x_g + \Delta_g;w), \varphi(x' + \Delta_s;w)),
\end{equation}
where $\left \| \cdot \right \|_{p}$ is the $\ell_{p}$-norm, $\epsilon$ and $\tau$ control the strength of perturbations, $\Delta_g$ and $\Delta_s$ are the data space perturbations for mitigating the approximation error in parameter space, \yonggang{and $x'$ denotes the perturbed version of source image}. In this way, we reduce the error in parameter space to  approximate the target model. 
\yonggang{
We solve the inner maximization problem by generating perturbations in the feature space, given a perturbed adversarial example. This step aims to mitigate the approximation error. The outer minimization problem is solved by finding adversarial perturbations in the input space, the same as the adversarial example generation. After iterative generation of perturbations, adversarial examples are generated by attacking a model with a reduced approximation error.
}

\section{Experiments}
\label{sec:Experiments}
In this section, we conduct extensive experiments to verify the power of lightweight black-box attacks augmenting with ETF.
%In Section~\ref{subsec:setup}, we introduce the experimental setups and implementation details. In Section~\ref{subsec:classification}, we show the effectiveness of our ETF on both normally and adversarially trained models in lightweight black-box attacks. In Section~\ref{Ablation}, we provide further analysis and ablation studies. 
 
\subsection{Experimental Setup}
\vspace{-1.5mm}
\label{subsec:setup}
% dataset 
% Surrogate model
\textbf{Models Architectures. }
All surrogate models are based on the ResNet-$18$~\cite{ResNet}. In the ordinary settings of the black-box attack, the general surrogate models are trained through the whole training set of ImageNet~\cite{ImageNet}. On the contrary, the lightweight surrogate models adopt only $1,000$ images randomly sampled from the validation set of ImageNet, considering that images in the training set are probably inaccessible in practice~\cite{nobox}. \yonggang{Results evaluated on the CIFAR10 dataset can be found in Appendix~\ref{cifar10results}}. Please refer to Table~\ref{Table_model} in the Appendix~\ref{Appendix:Detail} for the detailed information on the lightweight surrogate model and the general surrogate model. Regarding the target models, various model architectures are selected for full comparison, including,  VGG-$19$~\cite{VGG}, Inception v3~\cite{Inception}, ResNet-$152$~\cite{ResNet}, DenseNet~\cite{DenseNet}, SENet~\cite{SENet}, wide ResNet (WRN)~\cite{WRN} and MobileNet v2~\cite{MobileNet}.  
All these target models are well-trained on ImageNet.

\textbf{Implement Details. }   
The lightweight surrogate models are randomly initialized. The batch size is $128$, the epoch is $500$, and the initial learning rate is $0.4$, linearly decreasing to $0.008$. To better approximate the low-level feature of shallow layers, we refer to the self-supervision work~\cite{SimCLR} and apply various random augmentations to the data in each epoch for training.
To mount attacks, the classic methods, e.g., PGD~\cite{PGD}, MI~\cite{MI-FGSM}, DI~\cite{DI-FSGM}, and TI~\cite{TI-FSGM}, are applied to all the surrogate models. Unless otherwise specified, our attack is mounted on the first layer of ResNet-18 for lightweight surrogate models. 
\chsun{ Following previous works on intermediate-level attacks, the metric $d$ in Eq.~\ref{d_metric} of Shallow- in Table~\ref{Exp_main} is instantiated as the $\ell_2$-norm. Meanwhile, to maximize the potential of ETFs, we apply contrastive loss to ETF-($\cdot$) in Table~\ref{Exp_main}.   }
The Top-1 prediction accuracy is adopted as the evaluation metric, i.e., lower the classification accuracy means better attack success rate. All the experiments are run for $5$ individual trials with different random seeds. In our experiments, target models are evaluated using $\ell_\infty$-norm adversarial examples with maximum distortion $\epsilon=0.1$, results evaluated under mores settings, e.g., more strict constraints $\epsilon=0.05$, $\ell_2$-norm adversarial examples, and more test images 5,000 images, can be found in Appendix~\ref{Appendix:more experiment}.

\setlength{\tabcolsep}{4pt}
\begin{table}[tp]
\centering
\caption{The accuracy ($\%$) of $7$ normally trained target models evaluated on $1000$ adversarial examples generated by lightweight black-box attacks or existing black-box attacks, under $\epsilon \leq 0.1 $. Shallow-(PGD, MI, DI, TI) means applying PGD, MI, DI and TI to the shallow layers of the model. Deep-(PGD, MI, DI and TI) means applying PGD, MI, DI and TI to the model's output. EFT-(PGD, MI, DI and TI) means applying ETF combined with PGD, MI, DI or TI to the shallow layers.  
(The lower, the better)} 
\resizebox{\textwidth}{!}{
\begin{tabular}{llrrrrrrcr}
\toprule\noalign{\smallskip}
Model    & VGG19   & Inception  & RN152 & DenseNet & SENet & WRN   & MobileNet  & Average   \\
      &  \cite{VGG}
     &  v3\cite{Inception} 
     & \cite{ResNet} 
     & \cite{DenseNet} & \cite{SENet} 
     & \cite{WRN}   
     & v3\cite{MobileNet} &    \\
\noalign{\smallskip}
\toprule
      Clean &67.43     & 64.36         & 74.21 & 73.34    & 51.28 & 73.22 & 65.06     & 66.99  \\
%\midrule
Autoattack\cite{autoattack} & 0.00  & 0.00  & 0.00  & 0.00  & 0.00  & 0.00 & 0.00 & 0.00 \\
 \midrule
Deep-PGD          & 49.01±\small0.23 & 52.26±\small0.25 & 60.71±\small0.74 & 57.92±\small0.37 & 27.94±\small0.18 & 60.18±\small0.64 & 44.20±\small0.63 & 50.31±\small0.52 \\
 Deep-MI           & 38.92±\small0.43 & 42.37±\small0.37 & 49.53±\small0.49 & 49.06±\small0.89 & 19.44±\small0.75 & 49.11±\small0.82 & 33.46±\small0.80 & 40.69±\small0.96 \\
 Deep-DI           & 43.34±\small0.40 & 43.13±\small0.52 & 53.78±\small0.38 & 55.41±\small0.53 & 23.53±\small0.52 & 51.77±\small0.48 & 38.14±\small0.74 & 44.15±\small0.60 \\
 Deep-TI           & 49.46±\small0.52 & 49.64±\small0.27 & 58.89±\small0.71 & 58.75±\small0.30 & 26.19±\small0.16 & 56.31±\small0.58 & 44.02±\small0.46 & 49.03±\small0.51 \\ 
 Shallow-MI        & 22.62±\small0.25 & 30.83±\small0.48 & 34.05±\small0.27 & 35.74±\small0.76 & 12.31±\small0.41 & 29.98±\small0.65 & 17.72±\small0.31 & 26.17±\small0.56 \\
 Shallow-DI        & 22.14±\small0.39 & 29.78±\small0.17 & 35.51±\small0.33 & 35.79±\small0.61 & 8.99±\small0.42 & 30.61±\small0.88 & 16.88±\small0.47 & 25.67±\small0.55 \\
 Shallow-TI        &21.82±\small0.45  & 28.54±\small0.34 & 34.78±\small0.15 & 34.71±\small0.39 & 7.96±\small0.48 & 30.14±\small0.85 & 15.77±\small0.51 & 24.81±\small0.37 \\
 Shallow-PGD       & 22.93±\small0.33 & 31.07±\small0.58 & 34.71±\small0.67 & 36.20±\small0.87 & 13.08±\small0.36 & 32.16±\small0.66 & 16.65±\small0.54 & 26.69±\small0.49 \\ 
 ETF-PGD           & \textbf{14.11}±\small0.24 & 20.22±\small0.29 & 24.20±\small0.34 & 24.74±\small0.37 & 6.96±\small0.44  & \textbf{20.73}±\small0.28 & \textbf{10.66}±\small0.31 &  \textbf{17.37}±\small0.35  \\
ETF-MI             & 15.32±\small0.52 & 19.97±\small0.28 & 26.25±\small0.14 & 28.10±\small0.65 & 7.02±\small0.43 & 22.21±\small0.66 & 12.23±\small0.32 & 18.72±\small0.45 \\
ETF-DI             & 14.77±\small0.35 & 20.63±\small0.32 & 23.71±\small0.83 & 25.70±\small0.51 & 7.23±\small0.37 & 20.22±\small0.64 & 11.53±\small0.50 & 17.68±\small0.47 \\
ETF-TI             & 15.45±\small0.37 & \textbf{18.03}±\small0.34 & \textbf{22.63}±\small0.45 & \textbf{24.20}±\small0.68 & \textbf{6.94}±\small0.41 & 21.53±\small0.25 & 12.88±\small0.34 & 17.38±\small0.71 \\
\midrule 
 Deep*-PGD         & 12.43±\small0.51 & 28.15±\small0.43 & 16.54±\small0.49 & 12.61±\small0.22 & 7.09±\small0.32  & 13.33±\small0.54 & 9.64±\small0.28  & 14.25±\small0.37 \\
 Deep*-MI          & 11.77±\small0.75 & 25.14±\small0.56 & 18.10±\small0.64 & 13.72±\small0.34 & 4.26±\small0.35  & 14.61±\small0.37 & 8.30±\small0.37  & 13.70±\small0.68 \\
 Deep*-DI          & 7.61±\small0.41  & 18.17±\small0.45 & 8.23±\small0.33  & 9.90±\small0.57  & 6.66±\small0.34  & 9.72±\small0.42  & 7.91±\small0.46  & 9.74±\small0.55  \\
 Deep*\tnote{1}-TI   & 9.55±\small0.48  & 23.48±\small0.86 & 13.51±\small0.46 & 10.63±\small0.64 & 6.46±\small0.26  & 10.92±\small0.61 & 9.55±\small0.35  & 12.01±\small0.43 \\  
\bottomrule
	\end{tabular} }
	\begin{tablenotes} 
     \item[1]\vspace{-0.3em} \footnotesize Deep* refers to the attacks mounted in the model trained on the large-scale training data.  
   \end{tablenotes}
  \label{Exp_main}
\end{table}

\subsection{Main Results}
\label{subsec:classification}
\textbf{Results on normally trained models.} 
The results of classification accuracy are shown in Table~\ref{Exp_main}, which exhibits the evaluations for both lightweight and general surrogate models. The last four rows show the ordinary results where the general surrogate models are trained through all data in the training set. The black-box attacks are mounted based on the whole model, termed "Deep/Deep$^{*}$". As the size of the training data is reduced to $1,000$, this conventional method obviously loses its aggressiveness. Specifically, the averaged accuracies of 3-6 rows are among $40$-$50\%$, revealing limited attacking capability compared to $66.99\%$ of the clean accuracy.
It demonstrates the necessity of large amount of data for classic black-box attacks.
In rows 7-10, shallow layers without ETF are used to generate adversarial examples. The averaged accuracies of around $25\%$ illustrates the less dependence of shallow networks on the amount of data. 
In rows 11-14, attacks on shallow layers are further enhanced by ETF, decreasing the averaged accuracies to about $17\%$. The narrow gap between the proposed ETF and the general surrogate model indicates the potential capability of the black-box attack in the no-box settings.

\textbf{Results on Adversarially Trained Models.}
Here, we consider the models that are trained through adversarial training~\cite{PGD}, which is more persuasive to evaluate the effectiveness of attacking methods. Following~\cite{PGD}, the target models are trained on the whole training set of ImageNet. During adversarial training, the adversarial examples are generated under $\epsilon = 0,4/255,8/255$, where $\epsilon$ means the $\ell_{\infty}$-norm constraint. Table~\ref{adv_train} exhibits the attacking results mounted by ETF, black-box attacks (general surrogate models with all training data avaliable), and no-box attacks. For simplicity, only the results of PGD attacks are reported. As analysed above, the black-box attacks shows the best attacking results, while ETF only has a narrow gap with it. However, when $\epsilon = 4/255$ or $8/255$, ETF exceeds the black-box attacks ($29.13\%$/$26.14\%$ vs. $48.11\%$/$38.24\%$). It is interesting that black-box attacks are even worse than lightweight attacks. We believe that exploiting the phenomenon can bring something new, but we leave it as our future work due to the limited space.

\setlength{\tabcolsep}{4pt}
\begin{table}[t]
\centering
\caption{The accuracy ($\%$) of $7$ normally trained target models evaluated on $1000$ adversarial examples generated by no-box attacks and lightweight black-box attacks with different intermediate-level attacks. The best results are in \textbf{bold}. (The lower, the better)  } 
\setlength{\tabcolsep}{4pt}{
%\resizebox{\textwidth}{!}{
\begin{tabular}{lllllclcl}
\toprule\noalign{\smallskip}
Model    & VGG19   & Inception  & RN152 & DenseNet & SENet & WRN   & MobileNet  & Average   \\
       &  \cite{VGG}
     &  v3\cite{Inception} 
     & \cite{ResNet} 
     & \cite{DenseNet} & \cite{SENet} 
     & \cite{WRN}   
     & v3\cite{MobileNet} &    \\
\noalign{\smallskip}
\toprule
Clean        &67.43     & 64.36         & 74.21 & 73.34    & 51.28 & 73.22 & 65.06     & 66.99  \\
\midrule
No-box\cite{nobox}   & 18.74 & 33.68 & 34.72 & 26.06 & 42.36 & 33.16 & 16.34 & 29.29 \\
\midrule
ILA\cite{ILA}     & 20.13 & 28.01 & 35.72 & 35.14 & 9.16  & 29.97 & 14.31 & 24.63 \\
AA\cite{MSE-LOW} & 22.76 & 31.21 & 36.67 & 37.94 & 7.72  & 33.16 & 18.17 & 26.81 \\
ETF        & \textbf{14.11} & \textbf{20.22} & \textbf{24.20} & \textbf{24.74} & \textbf{6.96}  & \textbf{20.73} & \textbf{10.66} & \textbf{17.37} \\ 
\bottomrule
	\end{tabular} }
  \label{Exp_Nobox}
\end{table}

\subsection{Further Analysis and Ablation Study}
\label{Ablation}
\begin{wraptable}{r}{9cm}
\centering
\caption{ The performance of different attacks on the adversarial trained ResNet-$50$ \cite{robustness}. Therein, $\epsilon$ refers to the constraint $\ell_{\infty}$ in adversarial examples for adversarial training. The accuracy ($\%$) is evaluated on 1000 adversarial examples. $\epsilon = 0.1$ (the lower the better). White-box refers to Auto-Attack~\cite{autoattack}. }
\setlength{\tabcolsep}{1.0mm}{
%\resizebox{\textwidth}{!}{
\begin{tabular}{cccccc}
\toprule
\centering
Adv\_model  & Clean & ETF    & Black-box  & No-box  & White-box \\
  &   &  ours    &  ~\cite{PGD} & ~\cite{nobox} &  ~\cite{autoattack}\\
\midrule
\centering
$\epsilon=0/255$   & 69.43 & 16.97  & \textbf{8.20 }      & 24.53  & 0.00 \\
$\epsilon=4/255$ & 55.62  & \textbf{29.13}  & 48.11      & 39.62  &  0.00\\
$\epsilon=8/255$ & 41.68  &\textbf{ 26.14 } & 38.24      & 35.87  & 0.48  \\
\bottomrule
\end{tabular}}
\label{adv_train}
%\end{table}
\end{wraptable}
\textbf{Intermediate-Level Perturbations.} 
Since the lightweight black-box attacks are implemented mainly on intermediate features, two classic attacks mounted on intermediate features, i.e. ILA~\cite{ILA} and AA~\cite{MSE-LOW}, are selected for experiments. Comparing the ILA and AA in Table~\ref{Exp_Nobox} and Shallow-MI/DI/TI/PGD in Table~\ref{Exp_main}, the difference between their average accuracies is marginal. It indicates those intermediate-level feature perturbation has limited benefit to lightweight black-box attacks. 
The reason is that the intermediate-level attacks highly rely on the low-level features extracted by the lightweight model to generate adversarial examples and it can weaken the role of high-level information when attacking. The proposed ETF achieves better performance. On one hand, ETF does not rely on the features from high-level layers. On the other hand,
ETF introduces a min-max scheme to mitigate the adverse impact induced by the approximation error from the shallow layers.
% \textcolor{red}{ lightweight surrogate model are pushed to approximate a well-trained surrogate model. } 

\textbf{Comparison with No-box Attacks.} No-box attack is the first work to explore how to mount attacks in the no-box setting. Since Li et. al~\cite{nobox} stated that black-box attacks do not work in this setting, surrogate models are instantiated as auto-encoders rather than models for approximating target models. Namely, they design the pretext task to train $20$ auto-encoders per category to generate adversarial examples. However, the difference between the pretext tasks and the target tasks may significantly reduce the transferability of adversarial examples. In contrast, our method trains the lightweight surrogate model for tasks similar to or same as the task of target models. Hence, though only one sample per category is available, we achieve better performance than no-box attacks, as shown in Table~\ref{Exp_Nobox}.

% \begin{figure}[t]	
% 	\subfigure[]
% 	{
% 		\begin{minipage}{7cm} \label{fig:num}
% 			\centering          %子图居中
% 			\includegraphics[scale=0.22]{number_of_image.pdf} 
% 		\end{minipage}
% 	} 
% 	\subfigure[]
% 	{
% 		\begin{minipage}{7cm}  \label{fig:idx}
% 			\centering      %子图居中
% 			\includegraphics[scale=0.22]{layer.pdf}  
% 		\end{minipage}
% 	}
% 	%\caption{name of the figure} %  %大图名称
% 	%\label{fig:1}  %图片引用标记
% \caption{(a) How the lightweight attack performance of our approach varies with the number of images used for training the surrogate model. (b) The influence of low-level feature extraction at different layers of ResNet-$18$ on lightweight black-box attack performance.   (The lower, the better) }
% \vspace{-8mm}
% \label{numbers_and_layers}
% \end{figure}
\begin{figure*}[t] %t
\centering
\includegraphics[width=1\linewidth]{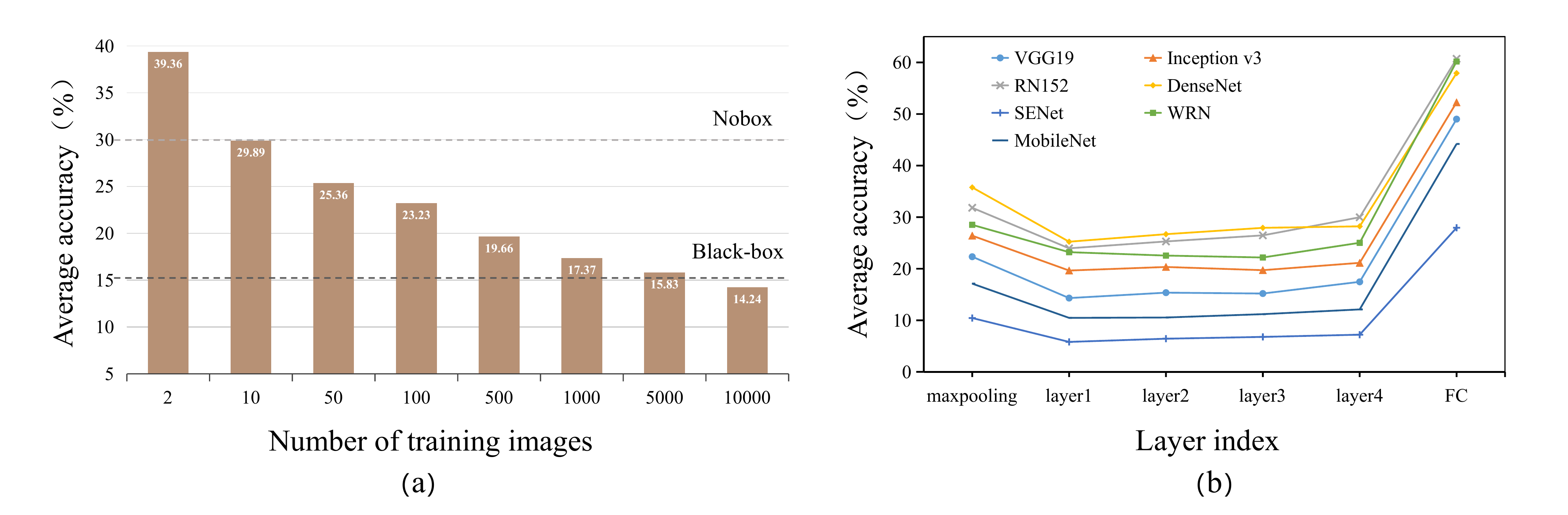}
\vspace{-8mm}
\caption{(a) How the lightweight attack performance of our approach varies with the number of images used for training the surrogate model. (b) The influence of low-level feature extraction at different layers of ResNet-$18$ on lightweight black-box attack performance.   (The lower, the better) }
\vspace{-2mm}
\label{fig:numbers_and_layers}
\end{figure*}

\textbf{Number of Training Samples.}
\label{subsec:number_of_samples}
One key difference between the lightweight black-box attack and existing black-box attacks mainly lies in the number of samples $n$ used for training surrogate models. Therefore, we further explore how lightweight attack performance varies with different number of samples, i.e., $n$. We conduct experiments on randomly selected $n$ images from the validation set of ImageNet. The experimental results are shown in Figure~\ref{fig:numbers_and_layers}. The average accuracy decreases as more samples are available, thus achieving better attacking results. More data makes it possible for the lightweight black-box attack to approximate the shallow layers of the target model more accurately. Especially the lightweight black-box attacks with $10,000$ test images can achieve the same performance as general black-box attacks which train the surrogate model on $1,200,000$ training images. Also, our method can still mount attacks even when only $2$ or $10$ images are available.

\textbf{Surrogate Model.} As analyzed above, we utilize the first few layers of the lightweight surrogate model to approximate the shallow layers of the target model to mount attacks. Hence, it would be interesting to study how the layers selection for lightweight models impacts the attack performance. We adopt different layers of ResNet-$18$ to generate adversarial examples utilizing ETF and attack the different target models. The results are summarized in Fig.~\ref{fig:numbers_and_layers}. It can be seen that low-level information in the first block of the model is sufficient for achieving promising attack performance. With more block information on the model, the performance of the attack does not improve.
In particular, the transferability of adversarial examples significantly decreases when we attack the fully-connected layer, verifying our analysis that the high-level layer of the lightweight model cannot approximate the target model well. More results for architecture selection can be found in Appendix~\ref{Appendix:more experiment}.

\begin{figure*}[t] %t
\centering
\includegraphics[width=1\linewidth]{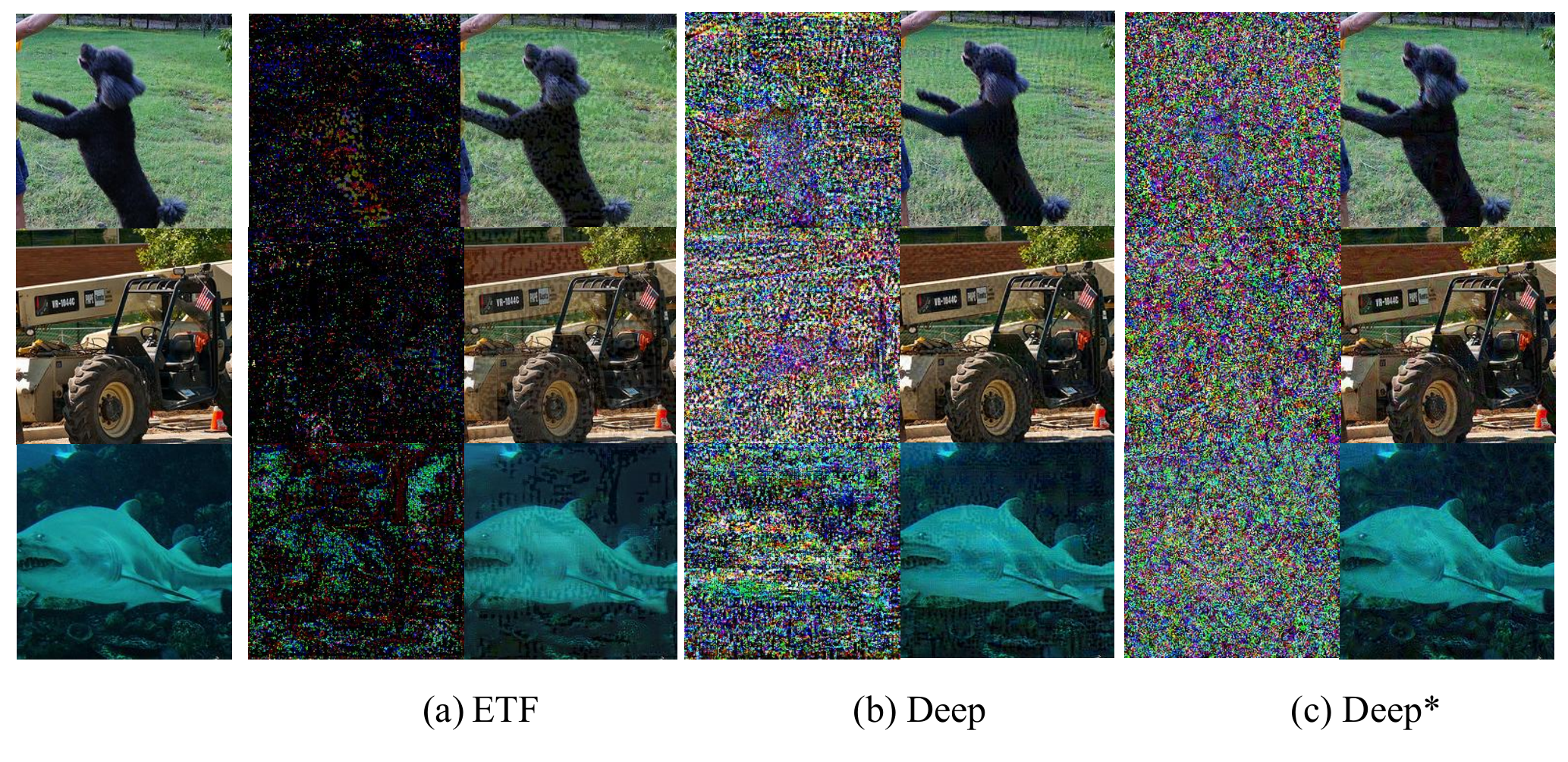}
\vspace{-8mm}
\caption{Adversarial examples crafted by: a)  ETF, b)  Deep, and c) Deep* attacks.}
\vspace{-2mm}
\label{Appendix:Adversarial examples}
\end{figure*}

\textbf{Adversarial Example Visualization.} To verify that adversarial examples are truly imperceptible, we provide visualization in Figure~\ref{Appendix:Adversarial examples}, where Deep*-PGD attack (using training images), Deep-PGD attack (using test images), and lightweight black-box attack are considered.

\setlength{\tabcolsep}{4pt}
\begin{table}[tp]
\centering
\caption{Model accuracy ($\%$) under lightweight black-box attacks under challenging scenarios, where supervision information or the in-distribution data are unavailable, named Unsupervised and OOD. } 
\resizebox{\textwidth}{!}{
\begin{tabular}{lrrrrrrrr}
\toprule\noalign{\smallskip}
Model    & VGG19   & Inception  & RN152 & DenseNet & SENet & WRN   & MobileNet  & Average   \\
       &  \cite{VGG}
     &  v3\cite{Inception} 
     & \cite{ResNet} 
     & \cite{DenseNet} & \cite{SENet} 
     & \cite{WRN}   
     & v3\cite{MobileNet} &    \\
\noalign{\smallskip}
\toprule
Clean        &67.43     & 64.36         & 74.21 & 73.34    & 51.28 & 73.22 & 65.06     & 66.99  \\
\midrule
Supervised  & 14.11 & 20.22 & \textbf{24.20} &  24.74  & 6.96  & \textbf{20.73} & 10.66 & 17.37 \\ 
Unsupervised & 15.54 &\textbf{ 19.16} & 26.27 & 23.75 & 7.66 & 22.79 & 11.43 & 18.08 \\
OOD &\textbf{ 6.13} & 21.72 & 25.44 & \textbf{21.89} & \textbf{5.02} & 24.33 &\textbf{ 7.16} &\textbf{ 15.96} \\ 
%Two         & 14.11 & 20.22 & 24.20 & 24.74 & 6.96  & 20.73 & 10.66 & 39.36 \\ 
%\midrule
\bottomrule
\vspace{-8mm}
	\end{tabular} }
  \label{Exp_OOD}
\end{table}

\section{Discussion}
\vspace{-3mm}
Though the no-box threat model provides a weak assumption, many realistic scenarios may be more complex. For example, adversaries have access to only data without supervision information due to security~\cite{security2018automatic} or privacy~\cite{survey} issues. In more extreme cases, e.g., autonomous vehicle~\cite{auto}, adversaries even have no access to the test samples from the distribution over which the model is trained. Instead, the adversaries have access to samples from a similar (but not exact) distribution (because the adversary knows the deployment environment). Hence, we consider two more challenging scenarios in which we provide solutions to mount lightweight black-box attacks.

\textbf{Insufficient supervision information.}
We design an experiment for mounting attacks using $1,000$ unlabeled samples randomly selected from the ImageNet validation set to explore the power of ETF under this challenging setting with unlabeled data.
Thanks to the framework of lightweight attack, it is feasible to train the unsupervised model on unlabeled data for attacks using Eq.~\ref{eq:train2}.  Specifically, we adopt contrastive learning~\cite{SimCLR} to train a lightweight surrogate model on the small-scale unlabeled dataset and generate adversarial examples by attacking the trained lightweight surrogate model. The performance is shown in Table~\ref{Exp_OOD}, where ETF still achieves similar performance without supervision information as that of the results under supervision. \yonggang{Intuitively, we can also employ other self-supervised learning method, e.g., rotation prediction, to train shallow layers, see details of experimental settings and results in Appendix~\ref{rotation}.}

\textbf{Consideration of Out-of-distribution.} We further study the possibility for mounting attacks under a more strict scenario, where only OOD (out-of-distribution, OOD) data are available. Specifically, we utilize the model pre-trained on OOD samples to mount lightweight black-box attacks. The rationality lies in that only low-level information is required to mount lightweight black-box attacks and the acquisition of low-level information. We load pre-trained ResNet-$50$ on STL-$10$~\cite{STL-10} dataset as the surrogate model. Without further fine-tuning, we experiment with two test in-distribution samples, and the results are recorded in Table~\ref{Exp_OOD}, named OOD. As we can see, the lightweight black-box attack with models trained on OOD data can achieve a surprising attack success rate, showing that black-box attacks enhanced with ETF can pose practical threats to deployed models. 

\vspace{-2mm}
\section{Conclusion}
\vspace{-2mm}
In this paper, we propose the conception of the lightweight black-box attack to reveal the potential risk of black-box attacks under the no-box threat model. To mount effective lightweight attacks, we find it crucial to leverage DNNs‘ shallow layers because they can learn similar features regardless of the number of samples. This is based on the empirical observations that the model trained with limited samples can approximate the shallow layers of the target models within acceptable errors, which can be used for crafting adversarial examples. Therefore, to further enhance the performance, we propose Error TransFormer (EFT) to decrease the approximation error by transforming the approximation error in the parameter space into the feature space. The experiments show that the proposed method achieves a surprising attack success rate under the no-box threat model using only one image per category, i.e., only $3\%$ lower than black-box attacks with complete training data of the target model.

\vspace{-2mm}
\section{Acknowledgements}
\vspace{-2mm}
This work was supported by NSFC No. 61872329 and No. 62222117,  and the Fundamental Research Funds for the Central Universities under contract WK3490000005. YZ and BH was supported by NSFC Young Scientists Fund No. 62006202 and Guangdong Basic and Applied Basic Research Foundation No. 2022A1515011652. TL was partially supported by Australian Research Council Projects DP180103424, DE-190101473,
IC-190100031, DP-220102121, and FT-220100318.

\bibliography{neurips_2022}

\clearpage

\appendix
\onecolumn

\section{ETF on Hidden Layers} \label{hiddern}
Before giving the details of applying ETF to hidden layers, we revisit the key identity:
\begin{small}
\begin{equation} \label{eq:trans_app}
\varphi(x; \left \{w^1 + w^1 A \right \} \cup \left \{ w \backslash w^1 \right \}) = 
g((w^1 + w^1 A) x; w \backslash w^1) =
g(w^1(x+Ax); w \backslash w^1) =
\varphi(x + A x ; w),
\end{equation} 
\end{small}
where $w^1$ stands for the first layer (convolution layer) parameters of model $\varphi$, $A$ is a transformation matrix used for perturbing $w^1$, $\left \{w^1 + w^1 A \right \} \cup \left \{ w \backslash w^1 \right \}$ means that the first layer's parameters are perturbed and the other layers' parameters keep the same, and $g$ is the function parameterized with $w \backslash w^1$ used for processing the first layer's outputs. Built upon Eq.~\ref{eq:trans_app},  we have $\varphi(x; \left \{w^1 + w^1 A \right \} \cup \left \{ w \backslash w^1 \right \}) = \varphi(x + A x ; w)$.

The model $\varphi$ can be decomposed: $\varphi=  g^l \circ \varphi^l$, where $\varphi^l(x)$ denotes the hidden feature at layer $l$. Then the output feature at layer $L$ is expressed as $\varphi^L(x;w) = g^l(w^l \varphi^l(x))$, where $w^l$ is the parameter used for processing the feature $\varphi^l(x)$. Then, we can apply the identity at layer $l$:
\begin{equation} \label{eq:trans_app_new}
\varphi^L \left(x;\left \{w^l + w^l A^l \right \} \cup \left \{ w \backslash w^l \right \}\right) = 
g^l\left((w^l + w^l A^l) \varphi^l(x)\right) =
g^l\left(w^l (A^l \varphi^l(x) + \varphi^l(x))\right),
\end{equation} 
where $A^l$ is the transformation matrix at layer $l$. According to Eq.~\ref{eq:trans_app_new}, we can transform a perturbation in the parameter space ($A w^l$) as the perturbation in the feature space ($A \varphi^l(x)$). Thus, lightweight black-box attacks with ETF generate adversarial examples as follows:
\begin{equation}
x_{adv} = \arg 
    \min_{\left \| x' - x\right \|_{p} \leq \epsilon} \  \max_{\left \| \Delta^l_s \right \|_{p} \leq \tau, \ 
    \left \| \Delta^l_g \right \|_{p} \leq \tau, l \in \left \{ 0,1, ..., L-1 \right \}}  
    d(\varphi(x_g, \cup_l \Delta^l_g;w), \varphi(x', \cup_l \Delta^l_s;w)),
\end{equation}
where $\Delta^l_s$ ($\Delta^l_g$) denotes the feature space perturbation ($l=0$ means the data space), $\cup_l \Delta^l_s$ ($\cup_l \Delta^l_g$) stands for all perturbations in the feature space, and $\varphi(x_g, \cup_l \Delta^l_g;w)$ ($\varphi(x_g, \cup_l \Delta^l_g;w)$) stands for the output feature with perturbed features, where the feature of source (guide) image at layer $l$ is perturbed by $\Delta^l_s$ ($\Delta^l_g$). Although we can perform the error transformation in the feature space, the features obtained using weights with approximation errors make it challenging. Specifically, we merely know that the input distribution (can be seen as the feature map) is not biased, but the feature map obtained using any weights will cause bias. Thus, we merely apply ETF to the input layer and leave further exploitation of applying it to hidden layers as future work.

\section{Additional Detail Description} \label{Appendix:Detail}
 
\subsection{Simplified Architecture}
The architecture of surrogate models is modified to avoid overfitting. Considering the limited amount of data, we employ a network with a small model capacity to instantiate the feature extractor. In particular, ResNet-$18$ is simplified by reducing the number of blocks in each layer, i.e., only one block is used in each layer of ResNet-$18$.

\subsection{Classification Ability of the Surrogate Model.} 

 \begin{wraptable}{r}{9cm}
 	\centering
\caption{The classification performance of the lightweight surrogate model and the general surrogate model.  } 
\begin{tabular}{ccccc} 
	\toprule
& \begin{tabular}[c]{@{}c@{}}Lightweight \\ surrogate model\end{tabular} & \begin{tabular}[c]{@{}c@{}}General \\ 
surrogate model\end{tabular} \\
\toprule
\begin{tabular}[c]{@{}c@{}}Number of \\ 
samples for training\end{tabular}   & 1000 & 1 200 000    \\
\midrule
\begin{tabular}[c]{@{}c@{}}Training \\ accuracy\end{tabular} & 96.38    &72.71   \\
\midrule
\begin{tabular}[c]{@{}c@{}}Test \\ accuracy\end{tabular}     & 2.36 & 63.24   \\
\bottomrule
\end{tabular}
\label{Table_model}
\end{wraptable}
To demonstrate that the lightweight black-box attack performance does not rely on the generalization of classification, we show the classification accuracy of the lightweight surrogate model used for ETF attacks in Table~\ref{Table_model}. To make the conclusion clearer, we also report the performance of a general surrogate model, which is trained on the training set of target models. We can see that the test accuracy of the lightweight surrogate model (about $2$) is drastically lower than that of the general model, bringing fresh air for black-box attacks. Specifically, the common sense in black-box attacks is that mounting attacks requires a surrogate model, which generalizes well on the test set. However, the experimental results in Table~\ref{Table_model} show that the lightweight surrogate model has poor classification performance, suggesting that the generalizability of surrogate models is not a necessary condition for performing black-box attacks.

\subsection{Approximation to more layers.}  Besides applying ETF in the first layer, we also conduct experiments that apply ETF to all layers except layer $L$. The results are reported in Table~\ref{conv1}. 
It can be seen that applying ETF to other layers can marginally promote the attack success rate. The phenomenon may result from the fact that the approximation error can accumulate with depth. Moreover, considering that applying ETF will cause more computational overhead, we merely give the results of applying ETF to the first layer in the main paper. 
%It can be seen that though more layers is used to approximate the target model by ETF, the attack performance is not improved greatly. 
%This may be because the mo  Besides, more approximation need more computation costs.   
%Therefore, in the main paper, we merely give the results of applying ETF to the first layer. 

\setlength{\tabcolsep}{4pt}
\begin{table}[thp]
\centering
\caption{ Apply ETF to all layers (except layer $L$) to further approximate the target model. This experiment is called "All" below. Similarly, "First" means only applying ETF to the first layer. (The lower, the better)   } 
\resizebox{\textwidth}{!}{
\begin{tabular}{lrrrrrrrr}
\toprule\noalign{\smallskip}
Model    & VGG19   & Inception  & RN152 & DenseNet & SENet & WRN   & MobileNet  & Average   \\
       &  \cite{VGG}
     &  v3\cite{Inception} 
     & \cite{ResNet} 
     & \cite{DenseNet} & \cite{SENet} 
     & \cite{WRN}   
     & v3\cite{MobileNet} &    \\
\noalign{\smallskip}
\toprule
Clean        &67.43     & 64.36         & 74.21 & 73.34    & 51.28 & 73.22 & 65.06     & 66.99  \\
\midrule
No-box\cite{nobox}   & 18.74 & 33.68 & 34.72 & 26.06 & 42.36 & 33.16 & 16.34 & 29.29 \\
\midrule
First  & 14.11 & 20.22 & 24.20  &  24.74  & 6.96  & 20.73  & 10.66 & 17.37 \\ 
All  & 13.33 & 20.21 & 22.66 & 25.49 & 5.40 & 21.74 & 9.88 & \textbf{16.95} \\
%Two         & 14.11 & 20.22 & 24.20 & 24.74 & 6.96  & 20.73 & 10.66 & 39.36 \\ 
%\midrule
\bottomrule
	\end{tabular}}
  \label{conv1}
\end{table}
\subsection{Hardware Configuration and Computation Costs.} 
We conduct experiments using GEFORCE RTX 2080 Ti, CPU AMD Ryzen 7 3700X @3.6 GHz. As merely 1,000 samples are required for the training of the lightweight surrogate model, the computational overhead is much less than the training of general surrogate models.

\yonggang{\section{Further Experiments on ImageNet}}
\label{Appendix:more experiment}

\yonggang{\textbf{Strict Constraint.}
We conduct experiments under smaller $\epsilon$, i.e., $\epsilon=0.05$. The results are given in Table~\ref{Appendix:eps}, demonstrating that ETF can generate powerful adversarial examples even with meeting more strict constraints, i.e., smaller $\epsilon$. 
}

\setlength{\tabcolsep}{4pt}
\begin{table}[thp]
\centering
\caption{ \yonggang{ The accuracy of 7 normally trained target models evaluated on 1,000 adversarial examples generated by lightweight black-box attacks or existing black-box attacks, under $\epsilon \leq 0.05$. The Shallow-(PGD, MI, DI, TI) mean applying PGD, MI, DI and TI to the shallow layers of the model. Deep-(PGD, MI, DI and TI) mean applying PGD, MI, DI and TI to the model’s output. EFT-(PGD, MI, DI and TI) mean applying ETF combined with PGD, MI, DI or TI to the shallow layers. }} 
\resizebox{\textwidth}{!}{
\begin{tabular}{lrrrrrrrr}
\toprule\noalign{\smallskip}
Model    & VGG19   & Inception  & RN152 & DenseNet & SENet & WRN   & MobileNet  & Average   \\
       &  \cite{VGG}
     &  v3\cite{Inception} 
     & \cite{ResNet} 
     & \cite{DenseNet} & \cite{SENet} 
     & \cite{WRN}   
     & v3\cite{MobileNet} &    \\
\noalign{\smallskip}
\toprule
Clean        &67.43     & 64.36         & 74.21 & 73.34    & 51.28 & 73.22 & 65.06     & 66.99  \\
Autoattack  & 0.00  & 0.20  & 0.00  & 0.00  & 0.00  & 0.10  & 0.00  & 0.04 \\
\midrule
Deep-PGD    & 61.14 & 63.05 & 65.78 & 62.31 & 34.50 & 68.17 & 56.65 & 58.8  \\
Shallow-PGD & 46.55 & 49.13 & 56.78 & 58.34 & 28.50 & 55.82 & 37.94 & 47.58 \\
ETF-PGD     & 41.76 & 46.74 & 48.55 & 50.79 & 24.68 & 53.11 & 32.65 & 42.61 \\
Deep*-PGD   & 16.23 & 36.71 & 25.36 & 24.62 & 18.16 & 31.42 & 13.34 & 23.69 \\

\bottomrule
	\end{tabular}} 
\label{Appendix:eps}
\end{table}

\yonggang{\textbf{More Validation Images.}
Besides the widely used setting on the number of samples, i.e., 1,000 images, we also evaluate different methods using more samples, i.e., 5,000 images, and report the results in Table~\ref{Appendix:5000}. The conclusion drawn from Table~\ref{Appendix:5000} is consistent with that drawn from Table~\ref{Exp_main}, e.g., EFT outperforms ``shallow'' attack methods, demonstrating that ETF can generate powerful adversarial examples under various scenarios.
}

\setlength{\tabcolsep}{4pt}
\begin{table}[thp]
\centering
\caption{\yonggang{ The accuracy of 7 normally trained target models evaluated on 5,000 adversarial examples generated by lightweight black-box attacks or existing black-box attacks, under $\epsilon \leq 0.1$. The Shallow-(PGD, MI, DI, TI) mean applying PGD, MI, DI and TI to the shallow layers of the model. Deep-(PGD, MI, DI and TI) mean applying PGD, MI, DI and TI to the model’s output. EFT-(PGD, MI, DI and TI) mean applying ETF combined with PGD, MI, DI or TI to the shallow layers. Auto-attack[23] is used for testing the robustness of the target models, so it adopts the white-box setting to mount the  target models. } }
\resizebox{\textwidth}{!}{
\begin{tabular}{lrrrrrrrr}
\toprule\noalign{\smallskip}
Model    & VGG19   & Inception  & RN152 & DenseNet & SENet & WRN   & MobileNet  & Average   \\
       &  \cite{VGG}
     &  v3\cite{Inception} 
     & \cite{ResNet} 
     & \cite{DenseNet} & \cite{SENet} 
     & \cite{WRN}   
     & v3\cite{MobileNet} &    \\
\noalign{\smallskip}
\toprule
Clean        &67.43     & 64.36         & 74.21 & 73.34    & 51.28 & 73.22 & 65.06     & 66.99  \\
 Autoattack\cite{autoattack} & 0.00  & 0.00  & 0.00  & 0.00  & 0.00  & 0.00 & 0.00 & 0.00 \\
\midrule
Deep-PGD    & 55.86          & 56.08          & 64.48          & 65.44          & 35.92         & 63.54          & 51.10         & 56.06          \\
Deep-MI     & 38.02          & 44.70          & 52.56          & 52.98          & 13.22         & 49.74          & 28.92         & 40.02          \\
Deep-DI     & 51.32          & 51.10          & 61.44          & 61.60          & 33.34         & 60.36          & 47.70         & 52.41          \\
Deep-TI     & 55.00          & 54.94          & 64.60          & 64.48          & 36.80         & 63.86          & 51.50         & 55.88          \\
Shallow-PGD & 19.42          & 25.12          & 31.04          & 31.70          & 9.28          & 29.16          & 16.64         & 23.19          \\
Shallow-MI  & 22.47          & 28.14          & 34.69          & 35.76          & 11.42         & 31.65          & 17.13         & 25.89          \\
Shallow-DI  & 19.68          & 24.62          & 30.26          & 32.17          & 10.02         & 28.24          & 16.08         & 23.01          \\
Shallow-TI  & 20.40          & 23.96          & 29.00          & 31.04          & 9.82          & 28.26          & 17.08         & 22.79          \\
ETF-PGD     & 13.56          & 17.66          & 23.68          & 24.60          & \textbf{4.54} & 20.68          & 9.42          & 16.31          \\
ETF-MI      & 15.94          & 20.32          & 26.28          & 26.74          & 5.52          & 22.72          & 9.70          & 18.17          \\
ETF-DI      & \textbf{13.16} & 25.72          & 22.32          & 22.76          & 4.68          & 19.84          & \textbf{8.58} & 15.29          \\
ETF-TI      & 13.30          & \textbf{14.60} & \textbf{20.48} & \textbf{22.38} & 5.22          & \textbf{19.06} & 9.50          & \textbf{14.93} \\
\midrule
Deep*-PGD   & 12.43          & 28.15          & 16.54          & 12.61          & 7.09          & 13.33          & 9.64          & 14.25          \\
Deep*-MI    & 11.77          & 25.14          & 18.10          & 13.72          & 4.26          & 14.61          & 8.30          & 13.70          \\
Deep*-DI    & 7.61           & 18.17          & 8.23           & 9.90           & 6.66          & 9.72           & 7.91          & 9.74           \\
Deep*-TI    & 9.55           & 23.48          & 13.51          & 10.63          & 6.46          & 10.92          & 9.55          & 12.01         \\
\bottomrule
	\end{tabular}} 
\label{Appendix:5000}
\end{table}

\yonggang{\textbf{$\ell_2$-norm Perturbation.}
We mainly conduct experiments with $\ell_\infty$ perturbation since it is widely adopted in many previous works~\citep{DI-FSGM,nes,TI-FSGM}. To further demonstrate the power of our ETF, we further evaluate different methods using $\ell_2$-norm perturbation. The results are reported in Table~\ref{Appendix:L2 experiment}, which further demonstrate the effectiveness of our proposal. Considering that $\ell_1$ and $\ell_0$ perturbations require careful design~\citep{l0,l1}, it is beyond the scope of this work, so we leave it as our future work. 
}

\setlength{\tabcolsep}{4pt}
\begin{table}[thp]
\centering
\caption{ \yonggang{ The classification accuracy evaluation on $\ell_2$-norm attacks. The experiment is conducted on the ImageNet validation. Following the previous work[25] about  $\ell_2$-norm  attacks, the maximum disturbance $\varepsilon$ is set to 16 $\sqrt[2]{N}$ where N is the dimension of input to attacks. } }
\resizebox{\textwidth}{!}{
\begin{tabular}{lrrrrrrrr}
\toprule\noalign{\smallskip}
Model    & VGG19   & Inception  & RN152 & DenseNet & SENet & WRN   & MobileNet  & Average   \\
       &  \cite{VGG}
     &  v3\cite{Inception} 
     & \cite{ResNet} 
     & \cite{DenseNet} & \cite{SENet} 
     & \cite{WRN}   
     & v3\cite{MobileNet} &    \\
\noalign{\smallskip}
\toprule
Clean        &67.43     & 64.36         & 74.21 & 73.34    & 51.28 & 73.22 & 65.06     & 66.99  \\
 Autoattack\cite{autoattack} & 0.00  & 0.00  & 0.00  & 0.00  & 0.00  & 0.00 & 0.00 & 0.00 \\
\midrule
Deep-PGD    & 37.73±\small0.31          & 42.75±\small0.34          & 51.04±\small0.77          & 51.96±\small0.62          & 17.48±\small0.34          & 50.61±\small0.49          & 31.07±\small0.55          & 40.38±\small0.54          \\
Deep-MI     & 40.40±\small0.44          & 45.02±\small0.51          & 54.53±\small0.46          & 54.13±\small0.53          & 17.59±\small0.47          & 53.22±\small0.63          & 32.47±\small0.41          & 42.48±\small0.55          \\
Deep-DI     & 38.73±\small0.53          & 38.63±\small0.49          & 50.34±\small0.35          & 48.79±\small0.48          & 17.66±\small0.43          & 47.53±\small0.57          & 27.34±\small0.33          & 38.43±\small0.47          \\
Deep-TI     & 37.89±\small0.23          & 37.86±\small0.38          & 46.52±\small0.46          & 45.62±\small0.31          & 18.54±\small0.44          & 46.44±\small0.37          & 30.75±\small0.52          & 37.66±\small0.46          \\
Shallow-PGD & 25.74±\small0.64          & 31.51±\small0.56          & 44.96±\small0.54          & 43.72±\small0.55          & 8.58±\small0.51           & 40.62±\small0.24          & 18.73±\small0.48          & 30.55±\small0.55          \\
Shallow-MI  & 37.46±\small0.94          & 42.28±\small0.87          & 51.56±\small0.79          & 50.77±\small0.63          & 16.58±\small0.67          & 52.06±\small0.86          & 28.02±\small0.74          & 39.82±\small0.67          \\
Shallow-DI  & 28.75±\small0.55          & 28.36±\small0.64          & 38.11±\small0.49          & 40.23±\small0.41          & 15.54±\small0.56          & 34.42±\small0.75          & 24.08±\small0.77          & 29.93±\small0.66          \\
Shallow-TI  & 30.28±\small0.36          & 31.55±\small0.40          & 37.69±\small0.39          & 38.44±\small0.48          & 14.52±\small0.48          & 35.26±\small0.27          & 23.54±\small0.19          & 30.18±\small0.38          \\
ETF-PGD     & \textbf{22.16±\small0.54} & 27.03±\small0.36          & 34.87±\small0.48          & 37.94±\small0.59          & \textbf{11.28±\small0.37} & 29.63±\small0.41          & \textbf{16.17±\small0.46} & 25.58±\small0.28          \\
ETF-MI      & 32.76±\small0.95          & 33.05±\small0.87          & 45.91±\small0.91          & 44.22±\small0.88          & 14.38±\small0.76          & 41.54±\small0.78          & 20.76±\small0.69          & 33.23±\small0.74          \\
ETF-DI      & 23.71±\small0.46          & \textbf{23.45±\small0.55} & \textbf{33.29±\small0.56} & \textbf{34.25±\small0.49} & 12.49±\small0.34          & \textbf{29.23±\small0.24} & 18.54±\small0.48          & \textbf{24.99±\small0.53} \\
ETF-TI      & 25.23±\small0.37          & 25.73±\small0.68          & 34.15±\small0.73          & 37.34±\small0.43          & 12.56±\small0.66          & 30.07±\small0.56          & 21.53±\small0.45          & 26.65±\small0.69          \\
\midrule
Deep*-PGD   & 7.65±\small0.42           & 22.88±\small0.34          & 11.44±\small0.12          & 11.23±\small0.44          & 4.56±\small0.71           & 9.69±\small0.78           & 8.03±\small0.46           & 10.78±\small0.45          \\
Deep*-MI    & 11.26±\small0.65          & 26.08±\small0.92          & 17.47±\small0.34          & 15.73±\small0.56          & 4.78±\small0.48           & 14.52±\small0.41          & 8.58±\small0.88           & 14.06±\small0.57          \\
Deep*-DI    & 1.04±\small0.34           & 11.04±\small0.54          & 1.68±\small0.48           & 1.39±\small0.51           & 0.77±\small0.32           & 3.01±\small0.29           & 0.56±\small0.41           & 2.78±\small0.42           \\
Deep*-TI    & 5.56±\small0.44           & 18.09±\small0.36          & 9.94±\small0.43           & 10.42±\small0.37          & 3.23±\small0.74           & 8.27±\small0.43           & 6.54±\small0.43           & 8.86±\small0.49          
\\
\bottomrule
	\end{tabular}} 
\label{Appendix:L2 experiment}
\end{table}

\yonggang{\textbf{Architecture Selection.}
We further exploit whether the architecture of surrogate models have significant impact on the performance of ETF. Specifically, we instantiate the shallow layers with different model architectures containing ResNet~\cite{ResNet}, VGG~\citep{VGG}, and SENet~\citep{SENet}. The results are reported in Table~\ref{Appendix:different model}, demonstrating that our EFT is powerful across various model architectures.}

\setlength{\tabcolsep}{4pt}
\begin{table}[thp]
\centering
\caption{ \yonggang{ Model accuracy under ETF attack with different architectures, containing SENet, VGG11, and ResNet18. } }
\resizebox{\textwidth}{!}{
\begin{tabular}{lrrrrrrrr}
\toprule\noalign{\smallskip}
Model    & VGG19   & Inception  & RN152 & DenseNet & SENet & WRN   & MobileNet  & Average   \\
       &  \cite{VGG}
     &  v3\cite{Inception} 
     & \cite{ResNet} 
     & \cite{DenseNet} & \cite{SENet} 
     & \cite{WRN}   
     & v3\cite{MobileNet} &    \\
\noalign{\smallskip}
\toprule
Clean        &67.43     & 64.36         & 74.21 & 73.34    & 51.28 & 73.22 & 65.06     & 66.99  \\
\midrule  
SENet~\cite{SENet}  & 23.44          & 28.42          & 35.07          & 31.64          & 6.73          & 28.19          & 11.80          & 23.61          \\
VGG11~\cite{VGG}  & 18.20          & 22.65          & 27.24          & 26.33          & \textbf{6.47} & 23.16          & 12.69          & 19.53          \\
Resnet~\cite{ResNet} & \textbf{14.11} & \textbf{20.22} & \textbf{24.20} & \textbf{24.74} & 6.96          & \textbf{20.73} & \textbf{10.66} & \textbf{17.37}\\
\bottomrule
	\end{tabular}} 
\label{Appendix:different model}
\end{table}

\yonggang{
\textbf{Heavy Data Augmentation.} We follow the empirical conclusion suggested in~\citep{low-level}, where heavy data augmentation is vital for training appropriate shallow models. Because appropriate shallow models are necessary for mounting lightweight black-box attacks, data augmentation plays a crucial role and is heavily used in our experiments. This is supported by results shown in Table~\ref{Appendix:augment}, where we report the performance of lightweight black-box attacks with and without data augmentation.
}

\setlength{\tabcolsep}{4pt}
\begin{table}[thp]
\centering
\caption{ \yonggang{ The impact of augmentation to ETF attacks. "No-Aug" means the effect of the attack on the ETF using the surrogate model without augmentation for training. This experiment is conducted on the ImageNet validation. The best results are in bold. } }
\resizebox{\textwidth}{!}{
\begin{tabular}{lrrrrrrrr}
\toprule\noalign{\smallskip}
Model    & VGG19   & Inception  & RN152 & DenseNet & SENet & WRN   & MobileNet  & Average   \\
       &  \cite{VGG}
     &  v3\cite{Inception} 
     & \cite{ResNet} 
     & \cite{DenseNet} & \cite{SENet} 
     & \cite{WRN}   
     & v3\cite{MobileNet} &    \\
\noalign{\smallskip}
\toprule
Clean        &67.43     & 64.36         & 74.21 & 73.34    & 51.28 & 73.22 & 65.06     & 66.99  \\
\midrule
No-Aug & 34.58          & 39.17          & 46.25          & 50.06          & 10.42         & 45.10          & 22.92          & 35.50          \\
Aug    & \textbf{14.11} & \textbf{20.22} & \textbf{24.20} & \textbf{24.74} & \textbf{6.96} & \textbf{20.73} & \textbf{10.66} & \textbf{17.37} \\
\bottomrule
	\end{tabular}} 
\label{Appendix:augment}
\end{table}

\textbf{Capacity to Evade Adversarial Detectors.} \chsun{Adversarial Detection~\citep{detector1,detector2,detector3} aims to distinguish adversarial examples from natural examples, which is also an effective way to test the robustness of adversarial attacks.} Therefore, we further exploit the capacity of ETF in evading adversarial example detectors. Specifically, we employ a detection method~\citep{detector1,detector2} to detect adversarial examples generated by different attack methods, e.g., FGSM~\citep{FGSM}, PGD~\citep{PGD}, BIM~\citep{MI-FGSM}, and ETF. All settings are the same as that used in the paper, and the results are reported in Table~\ref{Appendix:detection}. We can see that ETF performs better than the baselines, i.e., having a high probability of evading detection methods.

\setlength{\tabcolsep}{4pt}
\begin{table}[thp]
\centering
\caption{\yonggang{Performance of adversarial detection against four attacks, metric to evaluate the detection performance can be found in \cite{Mahalanobis,LID}. }} 
\setlength{\tabcolsep}{4pt}{
%\resizebox{\textwidth}{!}{
\begin{tabular}{lrrrrrrrr} 
\toprule\noalign{\smallskip}
\multicolumn{6}{l}{Mahalanobis~\cite{Mahalanobis}}                   \\
\noalign{\smallskip}
\toprule
Method    & TNR   & AUROC & DTACC & AUIN  & AUOUT \\ 
\midrule
BIM~\cite{BIM}        & 99.99 & 99.99 & 99.86 & 99.86 & 99.71 \\
FGSM~\cite{FGSM}      & 98.89 & 99.88 & 98.89 & 99.66 & 99.24 \\
Deep*-PGD~\cite{PGD} & 97.22 & 99.58 & 97.92 & 99.64 & 99.05 \\
ETF       & \textbf{96.67} & \textbf{98.73} & \textbf{96.94} & \textbf{98.75} & \textbf{97.98} \\
\toprule\noalign{\smallskip}
\multicolumn{6}{l}{LID\cite{LID}}                             \\
\toprule
Method    & TNR   & AUROC & DTACC & AUIN  & AUOUT \\
\midrule
Deep*-BIM~\cite{BIM}       & 99.99 & \textbf{98.81} & 98.33 & 99.77 & 99.33 \\
Deep*-FGSM~\cite{FGSM}      & 99.99 & 99.99 & 99.99 & 99.72 & 99.44 \\
Deep*-PGD~\cite{PGD} & 99.99 & 99.99 & 99.99 & 99.86 & 99.72 \\
ETF       & \textbf{97.78} & 99.58 & \textbf{97.22} & \textbf{99.51} & \textbf{98.68} \\
\bottomrule
	\end{tabular}} 
\label{Appendix:detection}
\end{table}

\section{Results on CIFAR10} \label{cifar10results}
\yonggang{We conduct the experiments on the CIFAR10 dataset, see Table~\ref{Appendix:cifar10 experiment}, and evaluate the robustness of models downloaded from RobustBench~\citep{robustbench}, see Table~\ref{Appendix:adv_cifar10}. The conclusion drawn from Table~\ref{Appendix:cifar10 experiment} and Table~\ref{Appendix:adv_cifar10} is consistent with that drawn from Table~\ref{Exp_main} evaluating on ImageNet dataset.}

\setlength{\tabcolsep}{4pt}
\begin{table}[thp]
\centering
\caption{ \yonggang{ Evaluate the performances of different attacks  on  CIFAR10. Here, experiments of "Deep-, Shallow-, ETF-" are conducted in the no-box threat model. "Deep*" means the black-box setting where the surrogate models are trained on the training data the same as the seven target models. "PGD~\cite{PGD}, MI~\cite{MI-FGSM}, DI~\cite{DI-FSGM}, TI~\cite{TI-FSGM}" is applied to the different settings and methods. Auto-attack[23] is used for testing the robustness of the target models, so it adopts the white-box setting to mount the seven target model. $\varepsilon \leq 0.1$ in $\ell_\infty$-norm. }} 
\resizebox{\textwidth}{!}{
\begin{tabular}{lrrrrrrrr}
\noalign{\smallskip}
\toprule
Model       & VGG19\cite{VGG}   & RN56\cite{ResNet}     & MobileNet\cite{MobileNet} & ShuffleNet\cite{SENet}  & Avg     \\
\toprule
clean       & 93.91 & 94.37 & 93.72   & 92.98    & 93.74 \\ 
Auto-attack~\cite{autoattack} & 0.00  & 0.00  & 0.00    & 0.00     & 0.00  \\
\midrule
Deep-PGD    & 59.45 ±\small 0.34  & 57.58 ±\small 0.46  & 45.21 ±\small 0.27  & 52.32 ±\small 0.37  & 53.64 ±\small 0.78  \\
Deep-MI     & 53.44 ±\small 0.75  & 52.17 ±\small 0.65  & 44.25 ±\small 0.34  & 49.80 ±\small 0.35  & 49.91 ±\small 0.58  \\
Deep-DI     & 60.24 ±\small 0.19  & 58.63 ±\small 0.34  & 47.67 ±\small 0.31  & 54.34 ±\small 0.62  & 55.22 ±\small 0.52  \\
Deep-TI     & 64.51 ±\small 0.38  & 59.85 ±\small 0.60  & 48.80 ±\small 0.59  & 56.88 ±\small 0.44  & 57.51 ±\small 0.42  \\
Shallow-PGD & 27.17 ±\small 0.74  & 31.06 ±\small 0.55  & 22.83 ±\small 0.66  & 28.14 ±\small 0.76  & 27.30 ±\small 0.81  \\
Shallow-MI  & 32.43 ±\small 0.98  & 36.42 ±\small 1.01  & 31.84 ±\small 0.79  & 30.76 ±\small 0.94  & 32.86 ±\small 0.94  \\
Shallow-DI  & 25.65 ±\small 0.56  & 30.27 ±\small 0.51  & 22.61 ±\small 0.38  & 27.22 ±\small 0.55  & 26.43 ±\small 0.45  \\
Shallow-TI  & 28.66 ±\small 0.45  & 31.35 ±\small 0.33  & 27.20 ±\small 0.44  & 29.48 ±\small 0.63  & 29.17 ±\small 0.56  \\
ETF-PGD     & 21.27 ±\small 0.27  & 25.85 ±\small 0.84   & \textbf{20.03} ±\small 0.65  & 22.37 ±\small 0.44  & 22.38 ±\small 0.53  \\
ETF-MI      & \textbf{20.75} ±\small 0.55  & \textbf{24.36} ±\small 0.35   & 20.51 ±\small 0.34  & \textbf{19.68} ±\small 0.23  & \textbf{21.32} ±\small 0.42  \\
ETF-DI      & 21.37 ±\small 0.37  & 26.46 ±\small 0.27   & 21.11 ±\small 0.69  & 23.14 ±\small 0.36  & 23.02 ±\small 0.55  \\
ETF-TI      & 25.48 ±\small 0.41  & 30.26 ±\small 0.23  & 23.37 ±\small 0.51  & 26.34 ±\small 0.25  & 26.36 ±\small 0.39  \\
\midrule
Deep*-PGD   & 4.63 ±\small 0.54   & 0.81 ±\small0.74    & 3.79 ±\small 0.28   & 3.21 ±\small 0.32   & 3.11 ±\small 0.47   \\
Deep*-MI    & 4.72 ±\small 0.20   & 0.96 ±\small 0.36   & 4.36 ±\small 0.12   & 3.78 ±\small 0.25   & 3.45 ±\small 0.33   \\
Deep*-DI    & 4.63 ±\small 0.17   & 0.81 ±\small 0.67   & 2.38 ±\small 0.53   & 3.34 ±\small 0.43   & 2.79 ±\small 0.47   \\
Deep*-TI    & 4.66 ±\small 0.18   & 0.84 ±\small 0.25   & 3.78 ±\small 0.46   & 3.67 ±\small 0.31   & 3.23 ±\small 0.32  \\ 
\bottomrule
	\end{tabular}} 
\label{Appendix:cifar10 experiment}
\end{table}

\setlength{\tabcolsep}{4pt}
\begin{table}[thp]
\centering
\caption{\yonggang{The attacks on the most robust models from CIFAR10 RobustBench. The robustness model is trained by the different adversarial defense method,$\varepsilon \leq 0.1$ in $\ell_\infty$-norm. }} 

\resizebox{\textwidth}{!}{
\begin{tabular}{lccccc}
\toprule\noalign{\smallskip}
  Model       & Gowal2021~\cite{Gowal2021} & Kang2021~\cite{Kang2021} & Pang2022~\cite{Pang2022} & Sehwag2021~\cite{Sehwag2021} & Avg     \\ 
\noalign{\smallskip} 
  \toprule
clean       & 89.00   & 92.00  & 87.50  & 86.50    & 88.75 \\
Auto-attack~\cite{autoattack} & 0.00  & 0.00  & 0.00    & 0.00     & 0.00  \\
\midrule
ETF-PGD     & 72.01   & 72.86  & 72.50  & 67.44    & 71.20 \\
Deep*-PGD   & 83.53   & 88.06  & 83.17  & 79.44    & 83.55 \\ 
\bottomrule
	\end{tabular}} 
\label{Appendix:adv_cifar10}
\end{table}

\section{Target Model Approximation Assumption } \label{assumption}
\yonggang{
Taking the first layer as an example, let $w^1_t$ and $w^1$ stand for the parameters of the target and surrogate models, respectively. In many practical scenarios, $w^1_t$ and $w^1$ usually have different dimensions, leading to intractable parameters’ discrepancy alleviation. Fortunately, we can find an appropriate low-rank approximation for parameters of deep neural networks~\citep{matrix1,matrix2,matrix3}. Specifically, we can approximate either $w^1_t$ or $w^1$ to make these two matrices have the same dimensions, so we can consider that the dimensions of the two models are the same. Consequently, we can find a transformation matrix $A$ such that the approximation error is minimized, i.e.,
$A =\arg\min_{\tilde{A}}|w^1_t -w^1-w^1\tilde{A}|_{F}$ , where $|\cdot|_{F}$ is the Frobenius norm. In this paper, we assume the approximation error is infinitesimal, i.e., $|w^1_t - w^1 - w^1A|_{F} = 0$. Then, we leverage $w^1$ and $A$ to represent the target model, i.e., $w^1_t = w^1 + w^1 A$.}

\section{Perturbation in Different Space } \label{perturbation}
\yonggang{
In the no-box setting, performing the min-max strategy in the feature space is more appropriate than the weight space optimization~\citep{awp} for the no-box threat model. This is because we know which perturbations are preferred in the feature space, e.g., towards features of guide images, but we have no idea about which perturbations are preferred in the weight space, i.e., no ``guide models'', which is supported by our experiments, see Table~\ref{Appendix:weight space}.}

\setlength{\tabcolsep}{4pt}
\begin{table}[thp]
\centering
\caption{\yonggang{The model accuracy under ETF attacks, where ``Feature\_space'' means the feature space perturbation and ``Weight\_space'' represents the min-max strategy in the weight space~\cite{awp}. This experiment is conducted on the ImageNet validation. The best results are in bold.} }

\resizebox{\textwidth}{!}{
\begin{tabular}{lrrrrrrrr}
\toprule\noalign{\smallskip}
Model    & VGG19   & Inception  & RN152 & DenseNet & SENet & WRN   & MobileNet  & Average   \\
       &  \cite{VGG}
     &  v3\cite{Inception} 
     & \cite{ResNet} 
     & \cite{DenseNet} & \cite{SENet} 
     & \cite{WRN}   
     & v3\cite{MobileNet} &    \\
\noalign{\smallskip}
\toprule
Clean        &67.43     & 64.36         & 74.21 & 73.34    & 51.28 & 73.22 & 65.06     & 66.99  \\
\midrule
Weight-space  & 29.43          & 32.44          & 40.11          & 41.88          & 10.12         & 35.41          & 19.27          & 29.81          \\
Feature-space & \textbf{14.11} & \textbf{20.22} & \textbf{24.20} & \textbf{24.74} & \textbf{6.96} & \textbf{20.73} & \textbf{10.66} & \textbf{17.37}  \\ 
\bottomrule
	\end{tabular}} 
\label{Appendix:weight space}
\end{table}

\section{Different Self-supervised Learning Approach } \label{rotation}
\yonggang{
It is straightforward that exploring different strategies to train the shallow model is exciting for further improvement of the performance of lightweight black-box attacks, as shallow layers play an important role in lightweight black-box attacks. Thus, we generate adversarial examples using EFT with shallow layers trained with a rotation prediction task~\citep{Rotations} and report the results in Table~\ref{Appendix:Rotation}. We can see that shallow layers trained with the rotation prediction task is slightly worse than using the contrastive strategy, but the performance can also reduce the model accuracy significantly.
}

\setlength{\tabcolsep}{4pt}
\begin{table}[thp]
\centering
\caption{  \yonggang{The model accuracy under ETF attacks, where ``rotation'' means that the shallow layers are trained using the rotation task and ``classification'' represents that shallow layers are trained with the classification task. This experiment is conducted on the ImageNet validation. The best results are in bold. }} 
\resizebox{\textwidth}{!}{
\begin{tabular}{lrrrrrrrr}
\toprule\noalign{\smallskip}
Model    & VGG19   & Inception  & RN152 & DenseNet & SENet & WRN   & MobileNet  & Average   \\
       &  \cite{VGG}
     &  v3\cite{Inception} 
     & \cite{ResNet} 
     & \cite{DenseNet} & \cite{SENet} 
     & \cite{WRN}   
     & v3\cite{MobileNet} &    \\
\noalign{\smallskip}
\toprule
Clean        &67.43     & 64.36         & 74.21 & 73.34    & 51.28 & 73.22 & 65.06     & 66.99  \\
\midrule
Rotation~\cite{Rotations}       & 19.07          & 21.79          & 27.30          & 28.85          & 7.66          & 23.94          & 12.51          & 20.16          \\
Classification & \textbf{14.11} & \textbf{20.22} & \textbf{24.20} & \textbf{24.74} & \textbf{6.96} & \textbf{20.73} & \textbf{10.66} & \textbf{17.37}\\
\bottomrule
	\end{tabular}} 
\label{Appendix:Rotation}
\end{table}

\section{Social Impact }
The motivation of this work is to provide an approach to evaluate the adversarial robustness in a more practical scenario, the no-box setting. Defenses can be assessed with fewer constraints through lightweight black-box attacks, i.e., without accessing training samples and any queries. We can develop defensive models robust against lightweight black-box attacks and attack algorithms to mislead deployed models. We believe the development of lightweight black-box attacks can help better access the robustness of deployed models and hope the proposed ETF can promote the development of corresponding defense methods.

\end{document}